%% file: main.tex
\documentclass[10pt,twocolumn,letterpaper]{article}

\usepackage[pagenumbers]{cvpr} % To force page numbers, e.g. for an arXiv version
\usepackage{multirow}

\usepackage{verbatim}
\usepackage{amsmath,amssymb,amsthm}
\usepackage{graphicx}
\usepackage{grffile}
\usepackage{epstopdf}
\usepackage{multirow}
 \usepackage{verbatim}
\usepackage{verbatim}
\usepackage{makecell}
\usepackage{pifont}
\usepackage{amssymb}
\usepackage{float}
\usepackage{graphicx}
\usepackage{multirow}

\usepackage{url}
\usepackage{textcomp}
\usepackage{epstopdf}
\usepackage{algorithm}
\usepackage{algpseudocode}
\usepackage{booktabs}
\usepackage{cases}
\usepackage{caption}
\usepackage{float}
\usepackage{threeparttable}
\usepackage{makecell}
\usepackage{placeins}

\input{preamble}

\definecolor{cvprblue}{rgb}{0.21,0.49,0.74}
\usepackage[pagebackref,breaklinks,colorlinks,citecolor=cvprblue]{hyperref}

\def\paperID{} % *** Enter the Paper ID here
\def\confName{}
\def\confYear{}

\title{Large Language Model Evaluated Stand-alone Attention-Assisted Graph Neural Network with Spatial and Structural Information Interaction for Precise Endoscopic Image Segmentation}

\author{\noindent\parbox{0.97\textwidth}{
\centering
Juntong Fan\textsuperscript{1,*},
Shuyi Fan\textsuperscript{2,*},
Debesh Jha\textsuperscript{3} \textit{IEEE Senior Member},
Changsheng Fang\textsuperscript{2}, \\
Tieyong Zeng\textsuperscript{1},
Hengyong Yu\textsuperscript{2,\dag} \textit{IEEE Fellow}, 
Dayang Wang\textsuperscript{2,\dag} \textit{IEEE Senior Member} 
\\ %\vspace{1.2em}
$^1$Department of Mathematics, The Chinese University of Hong Kong, Hong Kong. \\  %\qquad 
$^2$Department of Electronic and Computer Engineering, University of Massachusetts, Lowell, Lowell, MA, USA. \\%\qquad 
$^3$Department of Radiology, Northwestern University, Evanston, IL, US
\thanks{Juntong Fan and Shuyi Fan contribute equally,\\
\textsuperscript{\dag} Correspondence to Hengyong Yu (hengyong-yu@ieee.org) and Dayang Wang (dayang-wang@ieee.org)\\
}}
}

\begin{document}
\maketitle
% \input{sec/0_abstract}    
% \input{sec/1_intro}
% \input{sec/2_formatting}
% \input{sec/3_finalcopy}

% {
%     \small
%     \bibliographystyle{ieeenat_fullname}
%     \bibliography{main}
% }


\begin{abstract}
Accurate endoscopic images segmentation on the Polyps is critical for early colorectal cancer detection. However, this task remains challenging due to low contrast with surrounding mucosa, specular highlights, and indistinct boundaries. To address these challenges, we propose FOCUS-Med, which stands for Fusion of spatial and structural graph with attentiOnal features for Context-aware polyp segmentation in endoscopUS MEDical imaging. FOCUS-Med integrates a Dual Graph Convolutional Network (Dual-GCN) module to capture contextual spatial and topological structural dependencies. This graph-based representation enables the model to better distinguish polyps from background tissues by leveraging topological cues and spatial connectivity, which are often obscured in raw image intensities. It enhances the model’s ability to preserve boundaries and delineate complex shapes typical of polyps. In addition, a location fused stand-alone self-attention is employed to strengthen global context integration. To bridge the semantic gap between encoder-decoder layers, we incorporate a trainable weighted fast normalized fusion strategy for efficient multi-scale aggregation. Notably, we are the first to introduce the use of a Large Language Model (LLM) to provide expert-aligned qualitative evaluations of segmentation quality. Extensive experiments on public benchmarks demonstrate that FOCUS-Med achieves state-of-the-art performance across five key metrics, underscoring its effectiveness and clinical potential for AI-assisted colonoscopy.
\end{abstract}

\section{Introduction}
\label{sec:introduction}

\begin{figure}[htbp]  %bias
\centering
\includegraphics[width=\linewidth]{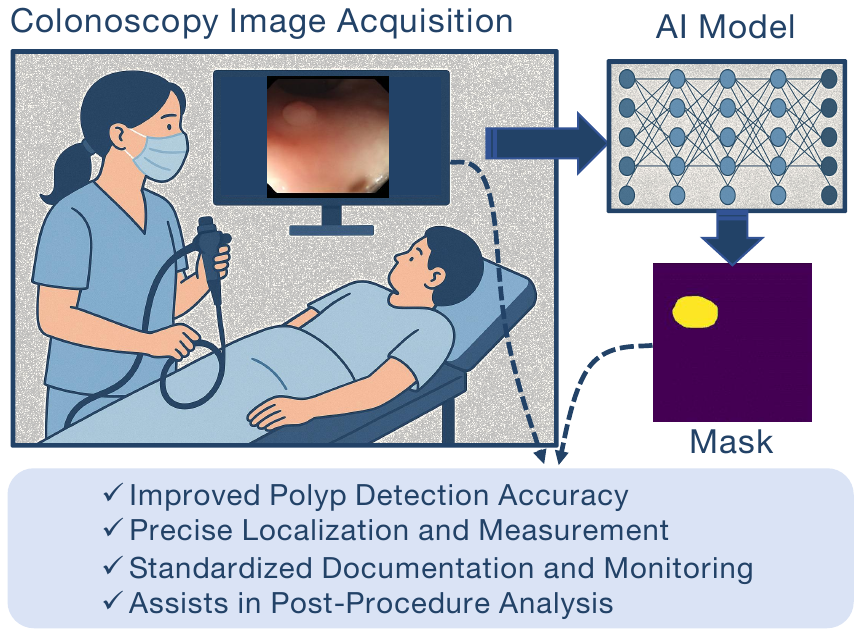} 
\caption{The clinical acquisition of Colonoscopy image and application of Polyps segmentation.} \label{overview}
\end{figure}

Colorectal cancer (CRC) is one of the leading causes of cancer-related deaths worldwide, with most cases developing from precancerous polyps over time \cite{granados2017colorectal}. Early and accurate detection of polyps \cite{shah2014biomarkers} during endoscopic colonoscopy is therefore critical for effective prevention and intervention. 
% Recent advancements in deep learning have enabled automated systems to assist gastroenterologists in detecting and segmenting polyps from endoscopic images. Among these, automatic polyp segmentation has emerged as a fundamental task due to its clinical relevance and potential to improve procedural outcomes.
As illustrated in Fig. \ref{overview}, clinically, precise segmentation of polyps enables several key benefits: (1) Improved detection accuracy, allowing even subtle or flat lesions to be recognized with higher confidence; (2) Precise localization and measurement, aiding in the assessment of polyp size, morphology, and resection margins; (3) Standardized documentation and monitoring, which ensures consistent records across clinical visits and facilitates training and audit trails; and (4) Assistance in post-procedure analysis, helping clinicians evaluate missed lesions or procedural errors retrospectively.

Despite its importance, polyp segmentation remains a challenging task \cite{dong2021polyp}. Polyps vary significantly in size, shape, color, and texture, and often exhibit visual similarity to surrounding mucosa. Poor lighting conditions, specular highlights, motion blur, and occlusion by bowel content further complicate the segmentation process. Small or flat polyps in particular are easily overlooked both by clinicians and automated systems.

\begin{figure*}[htbp]  %bias
		%	\addtocounter{figure}{+1}
\centering
\includegraphics[width=1\linewidth]{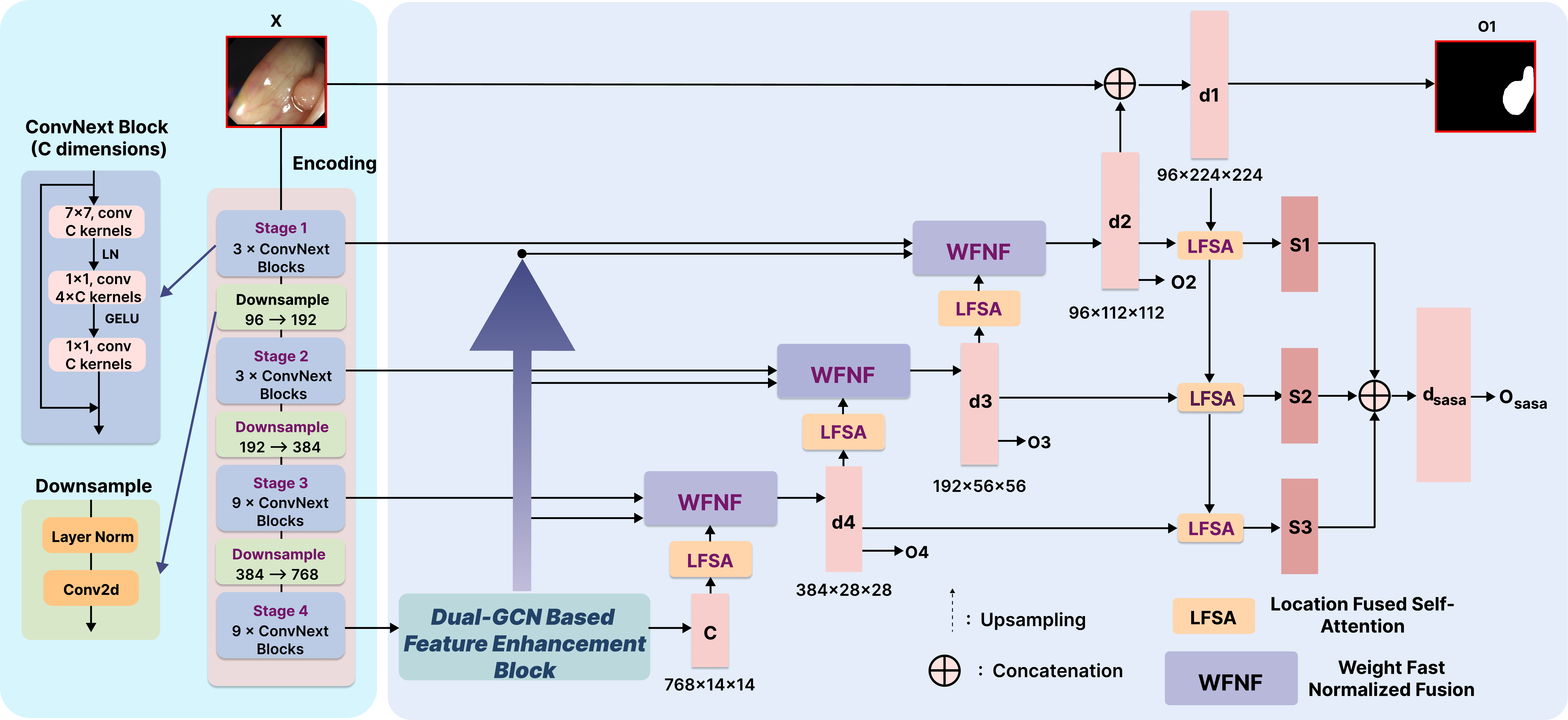} 
\caption{Overall frameworks of FOCUS-Med.} \label{model}
\end{figure*}

Inspired by the rapid advances of deep learning across diverse domains \cite{wu2023deep,wang2024lomae,han2024physics,jia2024enhancing,morovati2025patch}, numerous studies have explored specialized deep architectures for polyp segmentation. For instance, Fan et al. introduces a parallel reverse attention mechanism that combines region and boundary cues to refine polyp segmentation. By integrating a partial decoder and attention modules, it achieves high accuracy and high model efficiency \cite{47}. Liu et al. introduces a dual-branch framework that explicitly separates action from context for more accurate weakly-supervised temporal action localization. By distinguishing foreground-background and action-context snippets, it improves localization performance and significantly outperforms prior methods \cite{48}. Fang et al. proposes an embedding-unleashing framework for video polyp segmentation, which addresses challenges like low contrast and temporal variation by modeling appearance-level semantics. By integrating a proposal-generative network and an appearance-embedding network, the method enhances robustness to background noise and improves segmentation quality across frames \cite{fang2024embedding}.

While these deep learning–based approaches have shown promising results, existing methods still face limitations. Many models rely heavily on local texture cues and struggle with generalization across diverse datasets. Others lack explicit mechanisms for capturing long-range dependencies or contextual structures, which are essential for distinguishing polyps from background tissue in ambiguous cases.

In response to these challenges, we propose FOCUS-Med, a novel method that fuses spatial and structural graph representations with attentional features for polyp segmentation in endoscopic medical imaging. Our main contributions are as follows:

\begin{itemize}
\item We introduce the Dijkstra algorithm into graph neural networks for endoscopic image segmentation. This represents the first application of shortest-path-based graph attention in medical image segmentation, highlighting the novelty of our approach.
\item To enhance mask decoding capacity, we propose a location-fused self-attention module that strengthens the integration of global context, leading to improved segmentation performance.
\item To enable more effective multi-level feature integration, we develop a fast normalized fusion strategy with trainable weights. This module promotes semantic consistency and reduces information gaps across stages.
\item We are the first to explore the use of large language models (LLMs), such as GPT-4o, for evaluating segmentation quality. Experimental results further validate the strong performance of our proposed model.
\end{itemize}

\section{Related Work}
Please refer to Appendix for detailed related work. 

\section{Method}

% \subsection{Overview}

In this section, we introduce the proposed FOCUS-Med model in detail. As illustrated in Fig. \ref{model}, the encoder adopts ConvNeXt blocks \cite{liu2022convnet} for backbone feature extraction. To enhance deep feature interaction, we introduce a Dual Graph Convolutional Network-based Feature Enhancement Block (DBFEB) at the bottleneck. DBFEB builds spatial and structural graphs across channels to capture semantic dependencies, improving segmentation performance. In the decoder, we employ a location-fused stand-alone self-attention module to better focus on subtle and fine-grained features.

\subsection{Dual-GCN Based Feature Enhancement Block}
\subsubsection{Spatial and Structural Graphs Generation}

% Previous studies \cite{4} indicated that Dual-GCN can effectively extract spatial and structural information features from feature maps. 
Graph Convolutional Networks (GCNs) have demonstrated significant success in modeling complex, non-Euclidean relationships across a wide range of domains. Motivated by the irregular, dispersed morphology of polyps and their distribution across heterogeneous anatomical structures, we adopt a GCN-based strategy to enhance polyp segmentation. Specifically, we propose a feature enhancement block built upon Dual-GCN, which simultaneously captures fine-grained spatial details and long-range structural dependencies within endoscopic images. The spatial graph focuses on modeling localized pixel-level relations to preserve small structures, while the structural graph, constructed via shortest path attention, facilitates global context aggregation by encoding semantic interactions across distant anatomical regions.

% In recent years, GCN has been very popular in various fields due to its exceptional capacity in modeling the inter-relations between the neighboring neural nodes. In this context, we believe it is a perfect fit for the polyps segmentation problem since the whole structure of the polyps usually disperse across different kernel areas. The graph relation can further capture the spatial relation between different anatomical regions. Along this direction, we introduce a Dual-GCN based feature enhancement block to simultaneously extract both spatial and structural features from the endoscopic image. Specifically, this block consists of two parts: the first part is used to generate the spatial and structural graphs according to input data, and the other part is a feature extraction network to extract higher-level semantic information from the generated graphs. 

% \begin{figure}[htb]
% \begin{center}
% \includegraphics[width=.85\linewidth]{figures/gcn02.pdf}
% \caption{The process of generating spatial graph}
% \label{gcn02}
% \end{center}
% \end{figure} 

\begin{figure}[htb]
\begin{center}
\includegraphics[width=\linewidth]{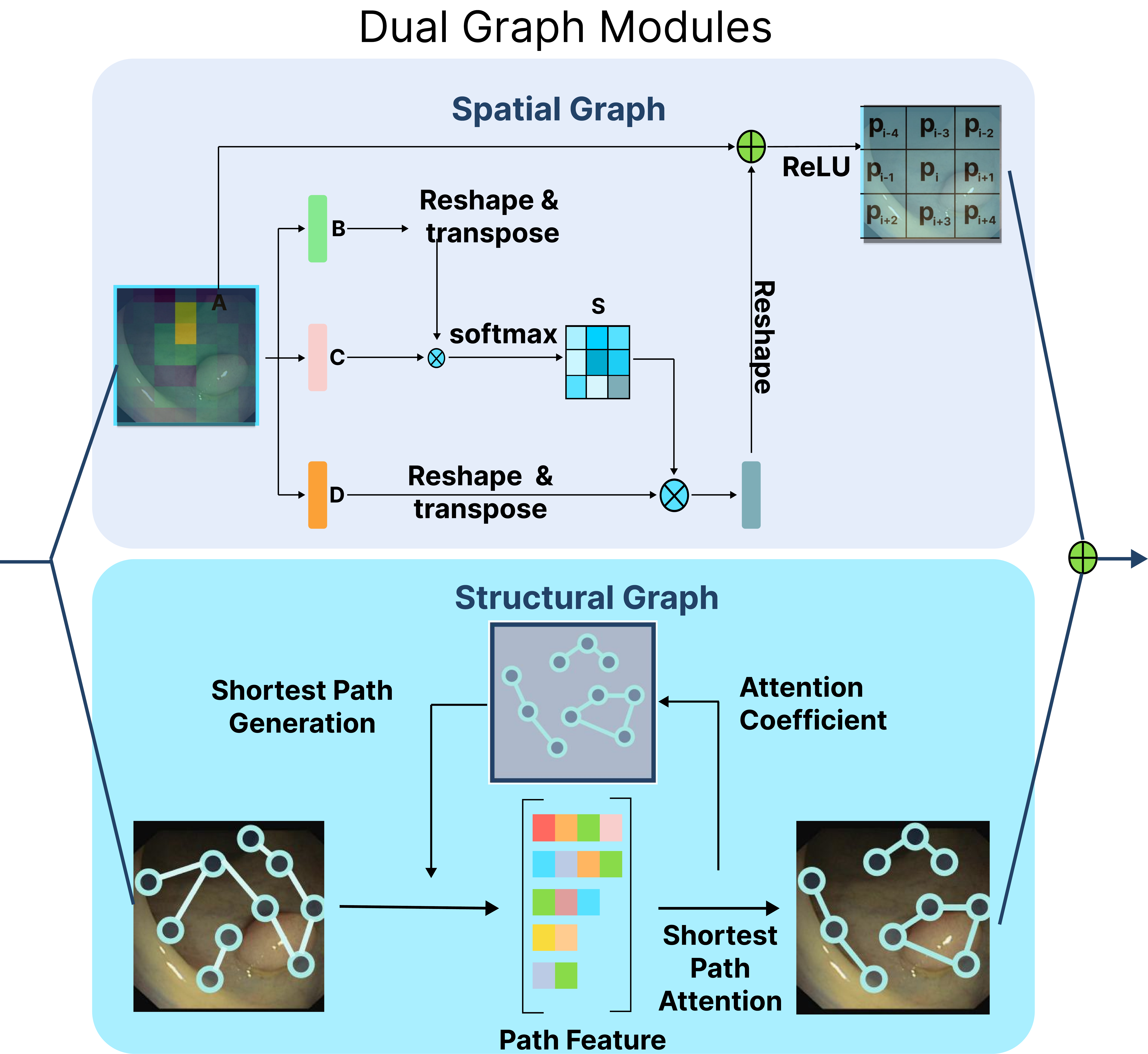}
\caption{The Dual-GCN enhancement block with contextual spatial and topological structural graphs. }
\label{gcn03}
\end{center}
\end{figure}

\textbf{Contextual Spatial Graph Construction}: To capture fine-grained local dependencies, we compute a pixelwise attention map based on feature similarity. Given input feature $A \in \mathbb{R}^{C \times H \times W}$, two $1 \times 1$ convolutions produce $B$ and $C$, reshaped to $\mathbb{R}^{C \times N}$ where $N = H \times W$. A third convolution yields $D \in \mathbb{R}^{C \times H \times W}$. The final spatially enhanced feature is computed as:

\begin{equation}
\hat{A} = \text{ReLU} \left( \left( \frac{\exp(B^\top C)}{\sum_{i=1}^N \exp(B^\top C)} \right) D + A \right),
\end{equation}

where the attention map captures pairwise similarity between pixels. This mechanism strengthens spatial focus on subtle patterns, improving the detection of small and irregular polyps.

\textbf{Topological Structural Graph Curation with Optimal Path Attention Mechanism}: To capture topological semantic dependencies in a structured and interpretable manner, we draw inspiration from recent advances in graph attention networks \cite{velivckovic2017graph, wang2019heterogeneous, yang2021spagan, vrahatis2024graph}. Specifically, we incorporate Dijkstra’s algorithm \cite{dijkstra1959note} to construct shortest-path-based graph structures that enhance the attention mechanism with interpretable topological priors for Polyps segmentation. Unlike conventional node-level attention that relies on first-order neighbors, our method enables each node to attend to high-order, semantically relevant nodes connected via low-cost shortest paths. As shown in Algorithm \ref{algorithm1}, this mechanism consists of two stages: (1) Optimal Path Formation with shortest length, (2) Multi-level path integration across length and attention heads. 

\textit{Step 1: Optimal Path Formation.}
Given a graph $G = (V, E)$ with node set $V$ and edge set $E$, we assign edge costs based on attention scores derived from previous layers. Let $\alpha_{ij}^{(k)}$ denote the attention coefficient from node $j$ to node $i$ in head $k$. The cost of edge $(i, j)$ is computed by averaging across all $K$ heads:

\begin{equation}
    \mathcal{W}_{ij} = \frac{1}{K} \sum_{k=1}^K \bar{\alpha}_{ij}^{(k)}.
\end{equation}

Since higher attention implies stronger semantic relevance, we treat $\mathcal{W}_{ij}$ as an inverse cost. These weights define a new attention-guided graph $\widetilde{G}$, over which we run Dijkstra’s algorithm to compute the minimum-cost paths from each node $s \in V$ to all other nodes $v \in V$.

Let $\text{path}_s(v)$ denote the sequence of nodes along the shortest path from $s$ to $v$, and $\text{cost}_s(v)$ its cumulative weight:

\begin{equation}
    \text{cost}_s(v) = \sum_{(i, j) \in \text{path}_s(v)} \mathcal{W}_{ij}.
\end{equation}

To limit the receptive field and remove trivial neighbors, we retain only paths where $2 \leq |\text{path}_s(v)| \leq C + 1$, with $C$ as the maximum path length. For computational efficiency and path diversity, we perform length-wise filtering. For each retained path length $c \in \{2, \dots, C+1\}$, we collect the set of paths of that length and select the top-$k$ lowest-cost paths for each node $s$, The filtered paths form a refined path set $\mathcal{N}_s^c$ for downstream aggregation. %where $k = \text{deg}(s) \cdot r$ and $r$ is a sampling ratio hyperparameter. 

\textit{Step 2: Multi-level Path Integration.}
To integrate features from sampled paths, we employ a two-level attention mechanism. At the first level, for each path $p \in \mathcal{N}_s^c$, we compute a path embedding $\phi(p)$ via mean pooling over node features along the path. The attention score between node $s$ and path $p$ in head $k$ is computed as:
% \, \Vert \,
\begin{equation}
    \alpha_{sp}^{(k)} = \text{softmax} \left( \langle a^{(k)}, h_s' \oplus \phi(p) \rangle \right),
\end{equation}

where $h_s'$ is the linearly transformed feature of node $s$. These are aggregated to form a feature representation $\ell_s^c$ for each path length $c$. At the second stage, cross-length attention is applied to fuse the representations across different values of $c$:

\begin{equation}
h_s^{\text{new}} = \sigma \left( \sum_{c=1}^C \text{softmax}\left( \langle b, h_s' \oplus \ell_s^c \rangle \right) \cdot \ell_s^c \right),
\end{equation}

where $\sigma$ is a non-linear activation and $\beta^c$ controls the importance of paths at different depths.

% \textit{Iterative Optimization.}
% The entire process is conducted iteratively. In the first iteration, edge weights $\mathcal{W}$ are initialized uniformly or using attention from a shallow network. After attention coefficients are updated through training, new edge weights are recomputed and shortest paths are regenerated accordingly. We empirically find that two iterations are sufficient to achieve stable performance. 

\subsection{Location Fused Stand-alone Self-Attention} \label{lfsa}
% Self-attention has been widely used in the transformer models \cite{8}. Inspired by different attention mechanisms, here, we proposed a novel self-attention block, named location fused stand-alone self-attention (LFSA) to the machine learning community. Particularly, LFSA is an independent plugin-and-play self-attention mechanism that enlarges the parameter space of the attention keys, thus potentially enhancing the model's performance. 

Self-attention has been extensively employed in transformer-based architectures due to its strong capacity for modeling long-range dependencies and capturing global contextual information \cite{8}. Motivated by ongoing advancements in attention mechanisms, we introduce a modified self-attention block named Location-Fused Stand-Alone Self-Attention (LFSA) to further capture irregular Polyps shape with more flexible design. Specifically, the process of building a new feature map using the LFSA module is expressed by:
\begin{equation}
\begin{split}
y_{ij} = \sum_{a,b\in N_k(i,j)} \text{Softmax}_{ab}(q^T_{ij}\cdot (\frac{\omega_{1}}{\varepsilon+\omega_{1}+\omega_{2}} \cdot k_{ab}\\
+ \frac{\omega_{2}}{\varepsilon+\omega_{1}+\omega_{2}} \cdot r_{a-i,b-j}))\cdot v_{ab}.
\end{split}
\end{equation}

\begin{algorithm}[H]
\caption{Shortest Path Attention via Dijkstra’s Algorithm}
\label{algorithm1}
\begin{algorithmic}[1]
\Require Graph $G = (V, E)$ with edge weights $\mathcal{W}$, node features $H = \{h_i\}$, max path length $C$, sampling ratio $r$, attention heads $K$
\Ensure Updated node representations $H = \{h_i^{\text{new}}\}$

\ForAll{$s \in V$} \Comment{Step 1: Optimal Path Formation (Dijkstra)}
    \ForAll{$v \in V$}
        \State dist[$v$] $\gets \infty$, path[$v$] $\gets$ empty list
    \EndFor
    \State dist[$s$] $\gets 0$, path[$s$] $\gets$ [$s$]
    \State $Q \gets$ min-priority queue of all nodes
    \While{$Q$ not empty}
        \State $u \gets$ node in $Q$ with smallest dist[$u$]
        \ForAll{neighbors $v$ of $u$}
            \State alt $\gets$ dist[$u$] + $\mathcal{W}[u][v]$
            \If{alt $<$ dist[$v$]}
                \State dist[$v$] $\gets$ alt
                \State path[$v$] $\gets$ path[$u$] + [$v$]
            \EndIf
        \EndFor
    \EndWhile

    \State $\mathcal{P}_s \gets$ all paths path[$v$] from $s$ where $2 \leq |\text{path}[v]| \leq C+1$ 

    \For{$c = 1$ to $C$}  %\Comment{Step 2: Path Sampling}
        \State $\mathcal{P}_s^c \gets$ paths of length $c$ in $\mathcal{P}_s$
        \State $k \gets$ degree($s$) $\cdot r$
        \State $\mathcal{N}_s^c \gets$ top-$k$ lowest-cost paths from $\mathcal{P}_s^c$
    \EndFor

    \For{$c = 1$ to $C$} \Comment{Step 2a: Integration over paths of same length}
        \For{$k = 1$ to $K$}
            \ForAll{$p \in \mathcal{N}_s^c$}
                \State $\phi(p) \gets$ mean-pooled features on path $p$
                \State $\alpha_{sp}^{(k)} \gets$ softmax attention between $h_s'$ and $\phi(p)$
            \EndFor
            \State $\ell_s^{c,k} \gets \sum \alpha_{sp}^{(k)} \cdot \phi(p)$
        \EndFor
        \State $\ell_s^c \gets$ concat($\ell_s^{c,1}, \ldots, \ell_s^{c,K}$)
    \EndFor

    \For{$c = 1$ to $C$} \Comment{Step 2b: Cross-length attention}
        \State $\beta^c \gets$ softmax attention between $h_s'$ and $\ell_s^c$
    \EndFor

    \State $h_s^{\text{new}} \gets \sigma \left( \sum \beta^c \cdot \ell_s^c \right)$
\EndFor

\Return $\{h_i^{\text{new}}\}$
\end{algorithmic}
\end{algorithm}

Here, $y$ represents a new feature map processed by LFSA model. $q,k,v$ represent queries, keys, and values, respectively. $q_{ij}=W_Q\cdot x_{ij}$, $k_{ab}=W_k\cdot x_{ab}$ and $v_{ab}=W_V\cdot x_{ab}$. $X$ represents an input image. $W_Q, W_K, W_V \in \mathbb{R}^{d_{out}\times d_{in}}$ are all learned transform. Different from the traditional self-attention, LFSA added two variables, row offset $a-i$ and column offset $b-j$, to express the relative distance of $ij$ to each position $ab \in N_k(i,j)$. $\omega_{1}$ and $\omega_{2}$ are both trainable parameters. $\varepsilon$ is a very small constant. Compared with traditional attention, this module is designed as a flexible, plug-and-play component that integrates spatial location information directly into the attention computation. By expanding the parameter space of attention keys through location fusion, LFSA enables more expressive feature representations, ultimately enhancing the model's ability to capture fine-grained spatial and contextual patterns for improved segmentation performance.

% The whole process is expressed as follows:
% \begin{equation}
% d_{lfsa} = S_1 \oplus S_2 \oplus S_3,
% \end{equation}
% Here, $d_{lfsa}$ is the final output and $S_i$ is calculated as:
% \begin{equation}
% S_i = \mathrm{LFSA}(d_i,d_{i+1}),
% \end{equation}
% \begin{equation}
% \mathrm{LFSA}(d_i,d_j) = y(q_i, k_i, v_j).
% \end{equation}

\subsection{Weighted Fast Normalized Fusion}

% In response to this challenge, Tan et al. \cite{9} introduced the fast normalized fusion (FNF) method, aiming for effective feature fusion. Nevertheless, the original method lacks consideration for the distinct contributions from different feature maps.
% The proposed DBFEB demonstrates proficiency in extracting rich spatial and structural features. However, a critical challenge arises in seamlessly fusing these features with corresponding maps during the feature inference process. As shown in Fig. \ref{model}. Three sources of information needs to be fused including prior feature, encoder shortcuts, and DBFEB features. Typical concatenation is overly expensive and addition risks information losing. To overcome this limitation, we furthermore present a Weighted Fast Normalized Fusion (WFNF) method, which takes into account the varying contributions from different feature maps. This enhancement results in a more versatile fusion process, allowing for the assignment of trainable weights to each feature map. The following equation outlines this refined procedure:
The proposed DBFEB effectively captures both spatial and structural representations. However, a key challenge lies in efficiently merging these features with the associated maps during the inference stage. As illustrated in Fig.~\ref{model}, three distinct sources must be integrated: the prior decoder feature, encoder skip connections, and DBFEB output. Simple concatenation can be computationally costly, while naive addition may cause loss of information. To address this, we introduce a Weighted Fast Normalized Fusion (WFNF) strategy, which assigns trainable weights to each input source, enabling efficient adaptive fusion based on their relative importance. The fusion process is formally defined as follows:

% The proposed DBFEB can extract rich spatial and structural features. However, one of the major challenges lies in effectively fusing these features with corresponding maps during the feature inference. Along this direction, Tan et al. \cite{9} introduced the fast normalized fusion (FNF) method for effective feature fusion. However, that method fails to account for the different contributions from different feature maps. To address this issue, we propose a modified fast normalized fusion (FNF) method which leverages the different feature map. It is a more efficient and faster fusion method that can assign trainable weights to each feature map. The following equation describes this procedure:
\begin{equation}
\begin{split}
O = & \sum_{k=1}^3 \frac{\omega_k}{\epsilon + \sum^6_{j=1}\omega_j}\cdot I_k 
 + \frac{\omega_4}{\epsilon + \sum^6_{j=1}\omega_j}\cdot \frac{I_1+I_2}{2} \\
& + \frac{\omega_5}{\epsilon + \sum^6_{j=1}\omega_j}\cdot \frac{I_1+I_3}{2} 
 + \frac{\omega_6}{\epsilon + \sum^6_{j=1}\omega_j}\cdot \frac{I_2+I_3}{2},
\end{split}
\end{equation}
In this formula, $\epsilon = 0.0001$ is a very small constant to avoid zero denominator. $I_k$ represents the $k^{th}$ feature map to be fused, and $I_k\in \mathbb{R}^{C\times H \times W}$.

% Moreover, compared with the fusion method with a softmax layer, fast normalized fusion can achieve comparable or better performance with $30\%$ faster training speed.

% \subsection{Loss Function}
% For model optimization, we utilize the weighted sum of BCE and Dice Losses, which can stabilize the training process and alleviate the imbalance between positive and negative samples. The specific formula is computed as follows:
% \begin{equation}
% \mathrm{Loss}(pred,Y)=\sum_{i=1}^4 \mathcal{L}(O_i, Y) + \mathcal{L}(O_{lfsa},Y),
% \end{equation}
% Wherein, $d_i$ represents the feature map generated by the $i_{th}$ layer of the model decoder. $Y$ is the label. $\mathcal{L}$ indicates the weighted sum of $BCE$ and $Dice$ losses, and the weight coefficient is set to $0.5$ by default. $O_{lfsa}$ denotes the fused results from the weighed fast normalized fusion process. 

% Different from the traditional loss function, we use the LFSA module to fuse $d_1$ with $d_2, d_3, d_4$ respectively, as shown in Figure \ref{structure}. The fusion output is noted as $O_{lfsa}$.

% The detailed mathematical formulas have been shown in section \ref{lfsa}. In the experiment, we find that the performance of the model will be improved obviously after adding $O_{lfsa}$.

\begin{figure}[htb]
\begin{center}
\includegraphics[width=\linewidth]{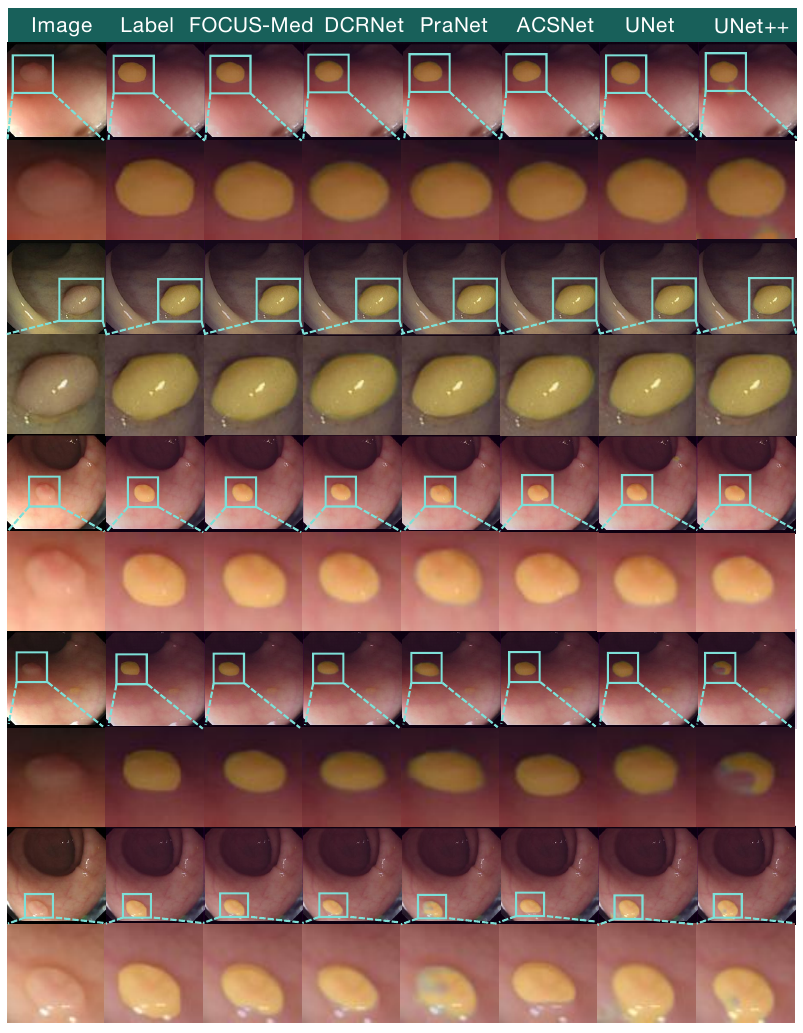}
\caption{Representative qualitative results of different models on different datasets, the results are the overlap of the masks with 80\% transparency on the original image. }
\label{Comparison}
\end{center}
\end{figure} 

% \begin{table*}[!htb]
% \begin{center}
% \begin{tabular}{ccccccc} %l(left)居左显示 r(right)居右显示 c居中显示
% \hline 
% Dataset & Model & Dice & IoU & MAE & Boundary\_F & S\_measure\\
% \hline
% \multirow{6}{*}{Endoscene} & U-Net  & $73.78\%$ & $66.54\%$ & $4.40\%$ & $68.78\%$ & $83.54\%$\\
% & U-Net++  & $72.88\%$ & $64.58\%$ & $4.50\%$ & $63.68\%$ & $82.41\%$\\
% & PraNet  & $81.73\%$ & $74.38\%$ & $3.50\%$ & $75.79\%$ & $88.00\%$\\
% & ACSNet & $85.15\%$ & $78.67\%$ & $3.00\%$ & $81.58\%$ & $90.54\%$\\
% & DCRNet  & $85.41\%$ & $78.86\%$ & $3.00\%$ & $83.20\%$ & $90.79\%$\\
% & Ours  & $\textbf{87.83\%}$ & $\textbf{81.65\%}$ & $\textbf{2.59\%}$ & $\textbf{83.29\%}$ & $\textbf{92.05\%}$\\
% \hline

% \multirow{6}{*}{Kvasir-SEG} & U-Net & $85.97\%$ & $78.70\%$ & $4.20\%$ & $73.13\%$ & $88.36\%$\\
% & U-Net++  & $84.16\%$ & $76.02\%$ & $5.20\%$ & $70.33\%$ & $87.17\%$\\
% & PraNet  & $89.20\%$ & $83.61\%$ & $3.10\%$ & $77.97\%$ & $90.96\%$\\
% & ACSNet  & $89.32\%$ & $83.83\%$ & $3.20\%$ & $79.04\%$ & $90.96\%$\\
% & DCRNet & $90.14\%$ & $84.44\%$ & $\textbf{2.90\%}$ & $\textbf{82.05\%}$ & $\textbf{91.49\%}$\\
% & Ours  & $\textbf{90.37\%}$ & $\textbf{84.46\%}$ & $3.20\%$ & $79.43\%$ & $91.43\%$\\
% \hline
% \end{tabular}
% \caption{Result Comparison on the Endoscene and Kvasir-SEG Dataset. The best results are bold-faced.}
% \label{table01}
% \end{center}
% \end{table*}

\begin{table*}[!htb]
\begin{center}
\setlength{\tabcolsep}{1mm}
{\fontsize{9pt}{10pt}\selectfont
\begin{tabular}{ccccccc}
\hline 
Dataset & Model & Dice & IoU & MAE & Boundary\_F & S\_measure\\
\hline

\multirow{6}{*}{Kvasir-SEG} & U-Net & $85.97\% \pm 0.02\%$ & $78.70\% \pm 0.03\%$ & $4.20\% \pm 0.02\%$ & $73.13\% \pm 0.04\%$ & $88.36\% \pm 0.01\%$\\
& U-Net++  & $84.16\% \pm 0.03\%$ & $76.02\% \pm 0.04\%$ & $5.20\% \pm 0.02\%$ & $70.33\% \pm 0.04\%$ & $87.17\% \pm 0.02\%$\\
& PraNet  & $89.20\% \pm 0.01\%$ & $83.61\% \pm 0.02\%$ & $3.10\% \pm 0.01\%$ & $77.97\% \pm 0.03\%$ & $90.96\% \pm 0.01\%$\\
& ACSNet  & $89.32\% \pm 0.02\%$ & $83.83\% \pm 0.01\%$ & $3.20\% \pm 0.01\%$ & $79.04\% \pm 0.02\%$ & $90.96\% \pm 0.01\%$\\
& DCRNet & $90.14\% \pm 0.01\%$ & $84.44\% \pm 0.01\%$ & $\textbf{2.90\%} \pm 0.01\%$ & $\textbf{82.05\%} \pm 0.01\%$ & $\textbf{91.49\%} \pm 0.01\%$\\
& FOCUS-Med*  & $\textbf{90.37\%} \pm 0.01\%$ & $\textbf{84.46\%} \pm 0.01\%$ & $3.20\% \pm 0.01\%$ & $79.43\% \pm 0.02\%$ & $91.43\% \pm 0.01\%$\\
\hline

\multirow{7}{*}{CVC-ClinicDB}
& UNET   & $66.79\% \pm 0.14\%$ & $60.02\% \pm 0.15\%$ & $3.07\% \pm 0.01\%$ & $59.36\% \pm 0.16\%$ & $80.40\% \pm 0.08\%$\\
& UNET++ & $76.13\% \pm 0.10\%$ & $68.20\% \pm 0.11\%$ & $2.77\% \pm 0.01\%$ & $65.02\% \pm 0.13\%$ & $85.29\% \pm 0.06\%$\\
& DCRNet & $89.24\% \pm 0.03\%$ & $83.17\% \pm 0.04\%$ & $0.76\% \pm 0.02\%$ & $80.42\% \pm 0.05\%$ & $93.66\% \pm 0.02\%$\\
& ACSNet & $87.94\% \pm 0.04\%$ & $82.96\% \pm 0.03\%$ & $3.91\% \pm 0.02\%$ & $84.59\% \pm 0.03\%$ & $90.56\% \pm 0.03\%$\\
& PraNet & $69.35\% \pm 0.13\%$ & $58.26\% \pm 0.16\%$ & $4.57\% \pm 0.02\%$ & $47.08\% \pm 0.21\%$ & $81.91\% \pm 0.07\%$\\
& UKAN   & $79.51\% \pm 0.08\%$ & $69.58\% \pm 0.10\%$ & $2.61\% \pm 0.01\%$ & $64.01\% \pm 0.13\%$ & $87.22\% \pm 0.06\%$\\
& FOCUS-Med* & $\textbf{94.74}\% \pm 0.02\%$ & $\textbf{90.05}\% \pm 0.02\%$ & $\textbf{0.51}\% \pm 0.01\%$ & $\textbf{90.81}\% \pm 0.02\%$ & $\textbf{96.46}\% \pm 0.01\%$\\
\hline

\multirow{6}{*}{Endoscene} & U-Net  & $73.78\% \pm 0.08\%$ & $66.54\% \pm 0.10\%$ & $4.40\% \pm 0.02\%$ & $68.78\% \pm 0.07\%$ & $83.54\% \pm 0.05\%$\\
& U-Net++  & $72.88\% \pm 0.08\%$ & $64.58\% \pm 0.10\%$ & $4.50\% \pm 0.02\%$ & $63.68\% \pm 0.10\%$ & $82.41\% \pm 0.05\%$\\
& PraNet  & $81.73\% \pm 0.05\%$ & $74.38\% \pm 0.05\%$ & $3.50\% \pm 0.01\%$ & $75.79\% \pm 0.04\%$ & $88.00\% \pm 0.02\%$\\
& ACSNet & $85.15\% \pm 0.03\%$ & $78.67\% \pm 0.03\%$ & $3.00\% \pm 0.01\%$ & $81.58\% \pm 0.01\%$ & $90.54\% \pm 0.01\%$\\
& DCRNet  & $85.41\% \pm 0.02\%$ & $78.86\% \pm 0.03\%$ & $3.00\% \pm 0.01\%$ & $83.20\% \pm 0.02\%$ & $90.79\% \pm 0.01\%$\\
& FOCUS-Med*  & $\textbf{87.83\%} \pm 0.02\%$ & $\textbf{81.65\%} \pm 0.02\%$ & $\textbf{2.59\%} \pm 0.01\%$ & $\textbf{83.29\%} \pm 0.02\%$ & $\textbf{92.05\%} \pm 0.01\%$\\
\hline

\multirow{7}{*}{CVC-ColonDB} 
& UNet        & $67.80\% \pm 0.08\%$ & $59.42\% \pm 0.10\%$ & $11.89\% \pm 0.03\%$ & $55.50\% \pm 0.07\%$ & $76.75\% \pm 0.05\%$ \\
& UNet++      & $67.96\% \pm 0.08\%$ & $58.98\% \pm 0.10\%$ & $10.01\% \pm 0.02\%$ & $54.38\% \pm 0.07\%$ & $79.01\% \pm 0.05\%$ \\
& DCRNet      & $93.56\% \pm 0.02\%$ & $88.08\% \pm 0.03\%$ & $\,\,\,0.80\% \pm 0.01\%$ & $86.66\% \pm 0.01\%$ & $95.59\% \pm 0.01\%$ \\
& ACSNet      & $94.86\% \pm 0.02\%$ & $90.37\% \pm 0.02\%$ & $\,\,\,0.61\% \pm 0.01\%$ & $90.87\% \pm 0.01\%$ & $95.83\% \pm 0.01\%$ \\
& PraNet      & $59.65\% \pm 0.08\%$ & $50.33\% \pm 0.10\%$ & $14.49\% \pm 0.04\%$ & $35.67\% \pm 0.07\%$ & $71.42\% \pm 0.05\%$ \\
& UKAN        & $87.20\% \pm 0.03\%$ & $77.77\% \pm 0.03\%$ & $\,\,\,1.86\% \pm 0.01\%$ & $68.18\% \pm 0.05\%$ & $90.16\% \pm 0.02\%$ \\
& FOCUS-Med*  & $\textbf{95.15\%} \pm 0.02\%$ & $\textbf{90.88\%} \pm 0.02\%$ & $\textbf{0.59\%} \pm 0.01\%$ & $\textbf{93.25\%} \pm 0.02\%$ & $\textbf{96.61\%} \pm 0.01\%$ \\
\hline
\hline

\end{tabular}
}
\caption{Result Comparison on different datasets. The best results are bold-faced.}
\label{table01}
\end{center}
\end{table*}

% \begin{table*}[!htb]
% \caption{Result Comparison on the Endoscene and Kvasir-SEG Dataset. The best results are bold-faced.}
% \begin{center}
% \scalebox{1}{
% \begin{tabular}{cccccccc} %l(left)居左显示 r(right)居右显示 c居中显示
% \hline 
% Dataset & Model & Year & Dice & IoU & MAE & Boundary\_F & S\_measure\\
% \hline
% \multirow{6}{*}{Endoscene} & U-Net\cite{10} & 2015 & $73.78\%$ & $66.54\%$ & $4.40\%$ & $68.78\%$ & $83.54\%$\\
% & U-Net++\cite{12} & 2019 & $72.88\%$ & $64.58\%$ & $4.50\%$ & $63.68\%$ & $82.41\%$\\
% & PraNet\cite{47} & 2020 & $81.73\%$ & $74.38\%$ & $3.50\%$ & $75.79\%$ & $88.00\%$\\
% & ACSNet\cite{48} & 2021 & $85.15\%$ & $78.67\%$ & $3.00\%$ & $81.58\%$ & $90.54\%$\\
% & DCRNet\cite{49} & 2022 & $85.41\%$ & $78.86\%$ & $3.00\%$ & $83.20\%$ & $90.79\%$\\
% & Ours & 2023 & $\textbf{87.83\%}$ & $\textbf{81.65\%}$ & $\textbf{2.59\%}$ & $\textbf{83.29\%}$ & $\textbf{92.05\%}$\\
% \hline

% \multirow{6}{*}{Kvasir-SEG} & U-Net\cite{10} & 2015 & $85.97\%$ & $78.70\%$ & $4.20\%$ & $73.13\%$ & $88.36\%$\\
% & U-Net++\cite{12} & 2019 & $84.16\%$ & $76.02\%$ & $5.20\%$ & $70.33\%$ & $87.17\%$\\
% & PraNet\cite{47} & 2020 & $89.20\%$ & $83.61\%$ & $3.10\%$ & $77.97\%$ & $90.96\%$\\
% & ACSNet\cite{48} & 2021 & $89.32\%$ & $83.83\%$ & $3.20\%$ & $79.04\%$ & $90.96\%$\\
% & DCRNet\cite{49} & 2022 & $90.14\%$ & $84.44\%$ & $\textbf{2.90\%}$ & $\textbf{82.05\%}$ & $\textbf{91.49\%}$\\
% & Ours & 2023 & $\textbf{90.37\%}$ & $\textbf{84.46\%}$ & $3.20\%$ & $79.43\%$ & $91.43\%$\\
% \hline
% \end{tabular}
% }
% \label{table01}
% \end{center}
% \end{table*}

\section{Experiments}
% \subsection{Data Description}
% All experiments are carried out on two polyps datasets: Endoscene and Kvasir-SEG. 
% \begin{enumerate}
%     \item Kvasir-SEG is a dataset of gastrointestinal polyp images\cite{25}. It is manually annotated by doctors and then verified by experienced gastroenterologists. Kvasir-SEG (file size is 46.2 MB) contains 1000 polyp images from Kvasir Dataset v2 and their corresponding truth values. The resolution of images contained in Kvasir-SEG ranges from $332\times487$ to $1920\times1072$ pixels.
%     \item Endoscene is an endoscopic image dataset generated by \cite{24}, which contains two subsets: CVC-ClinicDB DB and CVC-300. Endoscene contains a total of 912 images and corresponding pixel-level labels. The image size in the dataset is 612 images($384\times288$) and 300 images($574\times 500$), respectively.
% \end{enumerate}

% \subsection{Experimental Settings}
% All experiments were conducted on a single NVIDIA A40 (48GB) GPU with PyTorch framework. The input images were uniformly resized to 224×224 pixels for both training and testing. For training, we employed the Adam optimizer with an initial learning rate of $10^{-5}$ and a scheduled learning rate decay. The overall training process consisted of 300 epochs with a batch size of 2. 

\subsection{Comparative Experiments}

All experiments are run on Kvasir-SEG \cite{25}, CVC-ClinicDB \cite{bernal2015wm} , Endoscene \cite{24}, and CVC-ColonDB \cite{tajbakhsh2015automated} datasets. The test sets include 120 images from Kvasir-SEG, 30 from CVC-ClinicDB, 200 from CVC-ColonDB, and 80 from EndoScene. The proposed model is compared with state-of-the-art polyps segmentation models including U-Net \cite{10}, U-Net++ \cite{12}, PraNet \cite{47}, ACSNet \cite{48}, DCRNet \cite{49}, and UKAN \cite{li2025u}. All these models are implemented according to their officially disclosed codes. Particularly, five quantitative metrics are employed to fully assess the segmentation performances, which include Dice coefficient, intersection over union (IoU), mean absolute error (MAE), F1 score on boundary (Boundary\_F)\footnote{https://github.com/davisvideochallenge/davis2017-evaluation/blob/master/davis2017/metrics.py}, and structure measure (S\_measure) \cite{fan2017structure}.

\textbf{Qualitative analysis}. Qualitative results presented in Fig. \ref{Comparison} illustrate the efficacy of the studied methods in capturing the polyps' texture. While U-Net and U-Net++ tend to generate additional artifacts or distorted boundaries, more advanced models like PraNet, ACSNet, and DCRNet produce more accurate masks with oversmoothed boundaries. In contrast, the proposed model stands out by generating the most accurate polyps masks, capturing the clearest and most faithful polyps structure and boundaries compared to the counterparts. 

\begin{figure}[htb]
\begin{center}
\includegraphics[width=\linewidth]{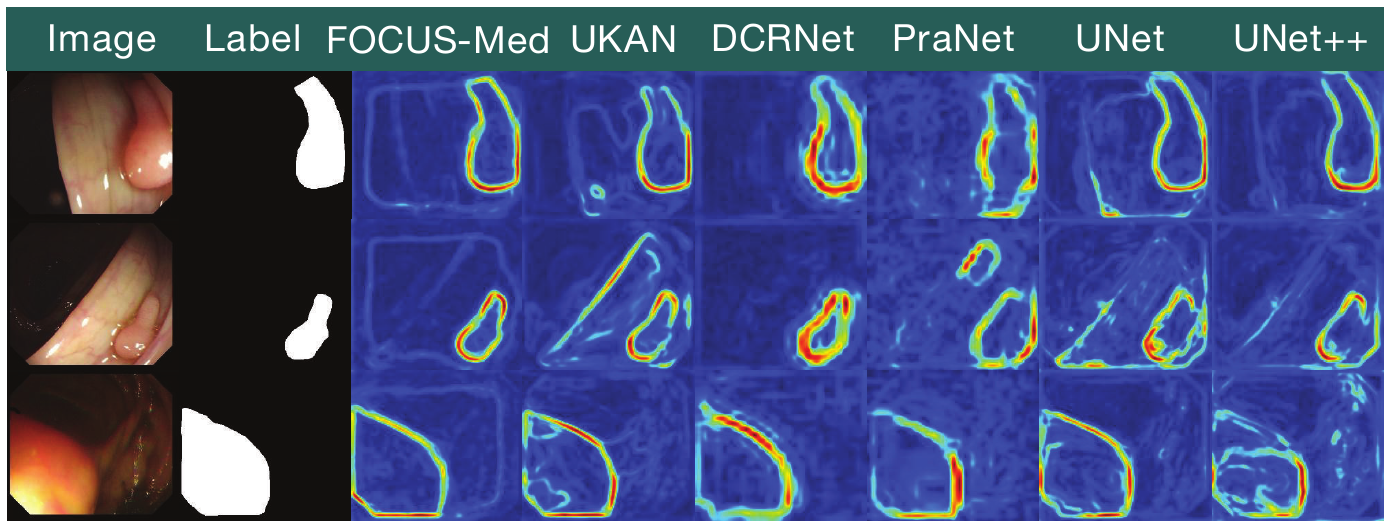}
\caption{The feature maps of different models on a repsentative images from CVC-ColonDB. }
\label{feature}
\end{center}
\end{figure}

\textbf{Feature map analysis}. The feature maps in Fig.~\ref{feature}, extracted from the last layer before softmax activation, further highlight the superiority of the proposed FOCUS-Med model in delineating polyp structures. Compared to other methods, FOCUS-Med consistently produces activation maps with sharper boundaries and stronger focus around the polyp regions, indicating higher confidence and spatial precision.

\textbf{Quantitative analysis}. The quantitative results are summarized in Table \ref{table01}. On the Endoscene, our model surpasses all other competitors in terms of all five indicators. It's noteworthy that the proposed model outperforms the second player with a large margin of more than 2\% on Dice and IoU scores. On Kvasir-SEG data, while the proposed model performs comparably to the competitors with respect to MAE, Boundary\_F, and S\_measure, it constantly delivers the highest Dice and IoU score. 

\textbf{Statistical Analysis}. To statistically validate the superiority of our proposed FOCUS-Med model, we conducted one-sided Wilcoxon signed-rank tests \cite{wilcoxon1945} on the Dice scores from the Kvasir-SEG dataset. The null hypothesis assumes that the compared state-of-the-art (SOTA) models outperform FOCUS-Med. The resulting p-values for each model are as follows: DCRNet (p=9.3e-4), UNet++ (p=9.3e-9), UNet (p=3.1e-8), ACSNet (p=9.1e-6), and ParNet (p=5.96e-7). As all p-values fall well below the conventional significance threshold (e.g., p $<$ 0.05), we reject the null hypothesis in each case, confirming that FOCUS-Med significantly outperforms existing models in Dice score on this dataset.

In summary, the proposed DSFNet can outperform the state-of-the-art models both qualitatively and quantitatively, showcasing its potential for clinical applications.

% \begin{comment}
% This shows that adding graph structure into the deep layer of neural network and using it to enhance the structural information and spatial information of images at the same time has a significant effect on training Endoscene data set, while other models do not involve the module of enhancing the structural information of images. 
% \end{comment}

% \begin{figure}[htb]
% \begin{center}
% \includegraphics[width=1\linewidth]{figures/loss.png}
% \caption{The Figure shows variation curves of loss during training for different models}
% \label{try}
% \end{center}
% \end{figure} 

% \subsection{Model Efficiency}

\subsection{Ablation Studies}
We conducted comprehensive ablation studies to evaluate the contribution of key components in our model, including the ConvNeXt backbone, the DBFEB, the LFSA module, and WFNF. The results demonstrate that each component plays a critical role in the overall performance of FOCUS-Med. Notably, the proposed WFNF method achieves the best segmentation performance in four out of five evaluation metrics. Furthermore, DBFEB consistently outperforms competing modules across all metrics, including Dice, IoU, MAE, Boundary\_F, and S\_measure. Please refer to the Appendix for detailed experimental results.

% \begin{figure}[hptb]
% \begin{center}
% \includegraphics[width=\linewidth]{figures/ablation02.pdf}
% \caption{The Figure shows qualitative results of DSFNet with different fusion methods}
% \label{ablation02}
% \end{center}
% \end{figure}

% \begin{comment}
% \begin{table*}[!htb]
% \caption{Attention Block Comparison}
% \centering
% \scalebox{0.8}{
% \begin{tabular}{ccccccc}
% \hline
% Dataset & \thead{Attention Block} & Dice & IoU & MAE & Boundary\_F & S\_measure\\
% \hline
% \multirow{3}{*}{Endoscene} & CCNet\cite{Huang_2019_ICCV} & $87.51\%$ & $81.04\%$ & $2.64\%$ & $82.43\%$ & $91.74\%$\\
% & CSNet\cite{10.1007/978-3-030-32239-7_80} & $87.16\%$ & $80.49\%$ & $2.64\%$ & $81.90\%$ & $91.75\%$\\
% & \thead{TransAttUnet\cite{chen2022transattunet}} & $87.53\%$ & $81.10\%$ & $2.61\%$ & $82.22\%$ & $91.65\%$\\
% & Ours & $\textbf{87.84\%}$ & $\textbf{81.65\%}$ & $\textbf{2.59\%}$ & $\textbf{83.29\%}$ & $\textbf{92.05\%}$\\
% \hline
% \end{tabular}
% }
% \label{table04}
% \end{table*}
% \end{comment}

% \FloatBarrier
\begin{figure*}[htb]
\begin{center}
\includegraphics[width=\linewidth]{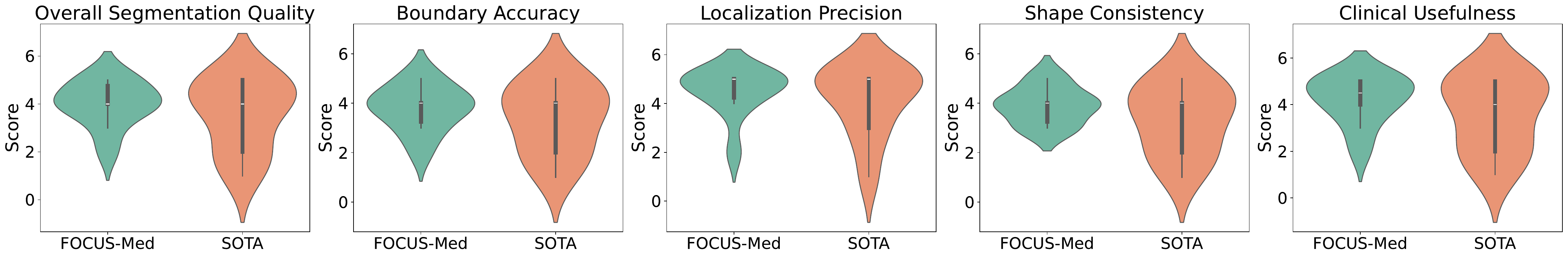}
\caption{The scores evaluated by LLM in terms of the five metrics. }
\label{llm}
\end{center}
\end{figure*}

\begin{table}[htbp]
\centering
\setlength{\tabcolsep}{1mm}
{\fontsize{9pt}{10pt}\selectfont
\begin{tabular}{c|cccc|cccc}
\hline
\multirow{2}{*}{Metric} & \multicolumn{4}{c|}{FOCUS-Med} & \multicolumn{4}{c}{SOTA} \\
 & Mean & Std & Q1 & Q3 & Mean & Std & Q1 & Q3 \\
\hline
OSQ & 4.10 & 0.99 & 4.00 & 5.00 & 3.56 & 1.51 & 2.00 & 5.00 \\
BA            & 3.80 & 0.92 & 3.25 & 4.00 & 3.33 & 1.41 & 2.00 & 4.00 \\
LP       & 4.50 & 0.97 & 4.25 & 5.00 & 4.11 & 1.45 & 3.00 & 5.00 \\
SC            & 3.90 & 0.74 & 3.25 & 4.00 & 3.33 & 1.41 & 2.00 & 4.00 \\
CU          & 4.20 & 1.03 & 4.00 & 5.00 & 3.56 & 1.59 & 2.00 & 5.00 \\
\hline
\end{tabular}
}
\caption{Descriptive statistics (Mean, Std, Q1, Q3) for clinical evaluation metrics of FOCUS-Med and SOTA.}
\label{reader}
\end{table}

\section{Segmentation Evaluation with LLM}
Recently, Large Language Models (LLMs) have been reported to exhibit emergent segmentation capabilities without explicit training \cite{sun2024training,wang2024llm,tang2025pre}. Motivated by this development, we introduce a novel, human-aligned evaluation framework that leverages LLMs to qualitatively assess polyp segmentation results. Rather than generating segmentations directly, we prompt LLMs to simulate clinical reasoning and provide structured judgments, offering complementary insights to traditional quantitative metrics such as Dice and IoU.

\textbf{Evaluation Setup.}
Specifically, we use GPT4o\footnote{https://chat.openai.com/chat} For each test sample, we provide the LLM with an input triplet consisting of (1) the original endoscopic image, (2) the predicted segmentation mask, and (3) the expert-annotated ground truth mask. The LLM is instructed to assess the segmentation performance across five clinically relevant dimensions using a 5-point Likert scale.

\textbf{LLM Evaluation Prompt.}
The LLM receives the following structured prompt for each image:

Prompt: \textit{You are reviewing the results of an AI-based system designed to segment polyps from endoscopic images. Each segmentation result includes the original image, the AI-predicted mask, and (if available) the expert ground truth mask. Your task is to evaluate the predicted segmentation across the five criteria below, 
Overall Segmentation Quality, Boundary Accuracy, Localization Precision, Shape Consistency, and Clinical Usefulness. using a 5-point Likert scale (1 = Poor, 5 = Excellent). } 

The metrics are specified as follows: 
\begin{itemize}
    \item Overall Segmentation Quality (OSQ): Does the predicted mask clearly and accurately highlight the polyp region?
    \item Boundary Accuracy (BA): Are the edges of the predicted mask well-aligned with the actual polyp boundary?
    \item Localization Precision (LP): Is the polyp mask centered and positioned correctly in the anatomical context?
    \item Shape Consistency (SC): Does the predicted mask capture the true morphology of the polyp (e.g., roundness, elongation)?
    \item Clinical Usefulness (CU): Would this segmentation meaningfully assist a gastroenterologist in real-world diagnosis or treatment?
\end{itemize}

\textbf{Results}. The performance of our proposed FOCUS-Med model was benchmarked against the best-performing SOTA method (DCR-Net) on the ColonDB dataset (Please find the Appendix for actual response of GPT4o to some example images). The violin plots in Fig. \ref{llm} visualize the distribution of LLM-evaluated scores across five key clinical metrics. Notably, for all metrics, FOCUS-Med displays distributions that are more peaked and concentrated around higher scores (centered at 4 or 5), while SOTA exhibits broader distributions with heavier tails toward lower ratings. This suggests that FOCUS-Med achieves not only higher average scores but also more consistent performance across different cases. The summary in Tab. \ref{reader} provides complementary evidence. FOCUS-Med consistently outperforms SOTA across all five metrics in terms of mean scores, with margins ranging from 0.3 to 0.7 points. The interquartile range (Q1–Q3) for FOCUS-Med is tightly clustered between 4 and 5 for all metrics, while DCR-Net shows wider spreads and lower Q1 values (as low as 2.0), indicating greater variability and the presence of underperforming samples. This is especially notable in Boundary Accuracy and Clinical Usefulness, where FOCUS-Med not only scores higher on average but also avoids the low tail that SOTA occasionally receives.

Taken together, these results indicate that FOCUS-Med not only achieves superior performance in terms of central tendency (mean and median), but also delivers more robust and reliable predictions across diverse cases, as evidenced by tighter variance and narrower interquartile ranges. These results are consistent with the earlier quantitative assessment presented in Tab. \ref{table01} and further reinforce the clinical promise of FOCUS-Med in polyp segmentation tasks.
% The Best SOTA model (DCRNet) is used to evaluate against over our FOCUS-Med model on ColonDB dataset. The violet map in Fig. \ref{llm} shows that the FOCUS-Med model tends to diliver performances more centered to high scores. Quantitative results in Table. \ref{reader} shows that for all these predefined metrics, the proposed FOCUS-Med outperforms SOTA, which aligns with the quantitative assessment showed in Tab. \ref{table01}. 

\begin{table}[t]
\centering
\setlength{\tabcolsep}{1mm}
{\fontsize{9pt}{10pt}\selectfont
\begin{tabular}{ccccc}
\hline
Model & Infer Time (ms) & FPS & Params (M) & Weights (MB) \\
\hline
UNet     & 72.30  & 13.83 & 34.53 & 131.81 \\
UNet++   & 74.95  & 13.34 & 36.63 & 139.85 \\
DCRNet   & 78.96  & 12.66 & 29.00 & 110.86 \\
ACSNet   & 96.19  & 10.40 & 29.45 & 112.58 \\
PraNet   & 95.43  & 10.48 & 32.55 & 124.72 \\
UKAN     & 81.10  & 12.33 & 25.36 & 97.08  \\
FOCUS-Med*   & 94.11  & 10.63 & 43.56 & 166.47 \\
\hline
\end{tabular}
}
\caption{Model complexity and inference speed comparison.}
\label{efficiency1}
\end{table}
\section{Discussion}
\textbf{Model Efficiency}. 
While segmentation accuracy is critical, real-time inference and model complexity are equally important for clinical deployment. As summarized in Tab. \ref{efficiency1}, we compare average inference time, frames per second (FPS), parameter count, and model size across several representative architectures. UNet and UNet++ achieve faster inference (72.30 ms and 74.95 ms per image, respectively) due to their shallow designs, but they exhibit limited segmentation accuracy compared to more recent approaches. In contrast, our proposed model, FOCUS-Med, achieves competitive inference performance (94.11 ms, 10.63 FPS) despite having a larger parameter count (43.56M) and weight size (166.47MB). This demonstrates its efficient design relative to its accuracy gains. Overall, FOCUS-Med maintains a comparable inference speed while delivering higher segmentation performance, illustrating its practical potential for real-time, high-precision medical applications.

\textbf{Base Module Investigation}. In addition to attention mechanisms, FOCUS-Med is primarily built on convolutional layers. Recently, architectures like Mamba \cite{mamba}, Kolmogorov-arnold networks (KAN) \cite{liu2024kan}, and diffusion-based models \cite{yang2023diffusion} have shown strong potential as backbones, offering enhanced sequence modeling and spatial reasoning. Our future work will explore integrating these advanced modules to further boost FOCUS-Med’s representational power and segmentation performance.

\begin{figure}[htb]
\begin{center}
\includegraphics[width=\linewidth]{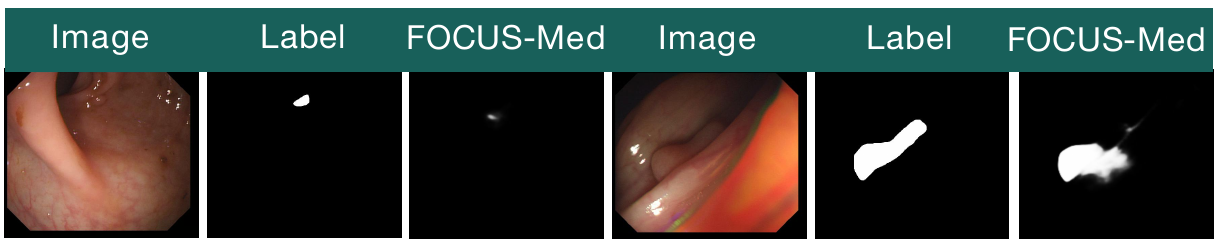}
\caption{The example of failing cases of FOCUS-Med from CVC-ColonDB. }
\label{fail}
\end{center}
\end{figure}

\textbf{Model Weakness}. 
In challenging cases where polyps are extremely small or visually resemble surrounding normal tissue, the proposed FOCUS-Med model sometimes struggles to accurately identify the true polyp regions as shown in Fig. \ref{fail}. In future work, we plan to enhance the model’s sensitivity by incorporating uncertainty-aware learning and external anatomical priors to better handle these ambiguous scenarios.

\section{Conclusion}
In this paper, we proposed FOCUS-Med, a novel framework for polyp segmentation that incorporates Dual-GCN for spatial–structural feature extraction, a location-fused self-attention module for global context refinement, and a weighted fast normalized fusion mechanism for multi-scale feature integration. To further validate segmentation quality beyond traditional metrics, we also explored the use of large language models (LLMs), such as GPT-4o, for expert-aligned qualitative assessment. Experimental results across multiple public datasets confirm that FOCUS-Med achieves state-of-the-art performance. In future work, we plan to extend this framework to other medical imaging tasks.

% In this paper, we proposed a Dual-GCN and location fused self-attention with weighted fast normalized fusion for polyps segmentation. Our model excels in capturing both spatial and structural features of polyp structures at a higher semantic level, thus enhancing the integration of global contextual information. Particularly, the model consists of three novel parts: first, the Dual-GCN feature enhancement module with the highlighted shortest path graph network was integrated into the bottleneck to enhance the feature extraction of local spatial and structural contextual information. Second, the LFSA module was incorporated to refine the global information. Finally, the weighted fast normalized fusion method with trainable weights was utilized to efficiently fuse diverse feature maps. As indicated by the experimental results on the public datasets, our proposed model demonstrates excellent segmentation performances and outperforms existing approaches in the gold standard metrics. In the future, we will explore the application of the proposed model to other medical applications. 

\bibliographystyle{unsrt} 
\bibliography{ref}

\end{document}

%% file: preamble.tex
\usepackage[dvipsnames]{xcolor}